\pdfoutput=1

\documentclass[11pt]{article}

\usepackage{acl}

\usepackage{times}
\usepackage{latexsym}

\usepackage[T1]{fontenc}

\usepackage[utf8]{inputenc}

\usepackage{microtype}

\usepackage{amsmath,amsfonts,bm}

\def\eqref#1{equation~\ref{#1}}

\def\1{\bm{1}}

\def\vu{{\bm{u}}}

\def\vx{{\bm{x}}}

\def\vz{{\bm{z}}}

\DeclareMathAlphabet{\mathsfit}{\encodingdefault}{\sfdefault}{m}{sl}
\SetMathAlphabet{\mathsfit}{bold}{\encodingdefault}{\sfdefault}{bx}{n}

\def\gG{{\mathcal{G}}}

\def\sB{{\mathbb{B}}}

\def\sD{{\mathbb{D}}}

\def\sL{{\mathbb{L}}}
\def\sM{{\mathbb{M}}}

\def\sR{{\mathbb{R}}}

\def\sT{{\mathbb{T}}}

\def\sV{{\mathbb{V}}}
\def\sW{{\mathbb{W}}}
\def\sX{{\mathbb{X}}}

\DeclareMathOperator*{\argmax}{argmax}

\usepackage{tikz}
\usepackage{xcolor}
\usepackage{booktabs}
\usepackage{amsmath}
\usepackage{comment}
\usepackage{multirow}
\usepackage{bbding}
\usepackage{graphicx}
\usepackage{subcaption}
\usepackage{adjustbox}
\usepackage[compact]{titlesec}
\usepackage{enumitem}
\usepackage[title]{appendix}

\newcommand{\RM}[1]{{\color{cyan}{RM: #1}}}

\newcommand{\bcr}{{\color{blue}{(BC-refs)}}}
\definecolor{mydarkgreen}{rgb}{0,0.5,0}

\newcommand{\Shape}{\textsc{Shape}}
\newcommand{\Thing}{\textsc{Thing}}

\title{Concept-Best-Matching:\\ Evaluating Compositionality in Emergent Communication}

\author{Boaz Carmeli, Yonatan Belinkov, Ron Meir \\
Technion -- Israel Institute of Technology \\
\texttt{boaz.carmeli@campus.technion.ac.il},\\ 
\texttt{belinkov@technion.ac.il}\\
\texttt{rmeir@ee.technion.ac.il}
}

\begin{document}
\maketitle
\begin{abstract}
Artificial agents that learn to communicate in order to accomplish a given task acquire communication protocols that are typically opaque to a human. 
A large body of work has attempted to evaluate the emergent communication via various evaluation measures, with  \emph{compositionality} featuring as a prominent desired trait.   
However, current evaluation procedures do not directly expose the compositionality of the emergent communication. 
We propose a procedure to assess the compositionality of emergent communication by finding the best-match between emerged words and natural language concepts.
The best-match algorithm provides both a global score and a translation-map from emergent words to natural language concepts. To the best of our knowledge, it is the first time that such direct and interpretable mapping between emergent words and human concepts is provided.

\end{abstract}

\section{Introduction}

Artificial agents that learn to communicate for accomplishing a given task acquire communication protocols. 
In the common setting, a sender observes a set of objects and sends a message to a receiver, which then needs to identify the correct target objects out of a set of distractors. 
The two agents learn a communication protocol over discrete vocabulary of atoms, termed ``words''. 
The emergent communication (EC) is typically opaque to a human. 
As a result, a large body of work has attempted to characterize the emergent communication in light of natural language (NL) traits, such as compositionality \citep{hupkes2020compositionality, chaabouni2020compositionality}, systematic generalization \citep{vani2021iterated}, pragmatism \citep{andreas2016reasoning, zaslavsky2020rate}, and more.

\begin{figure}[t]
\centering
\resizebox{8cm}{!} {\includegraphics{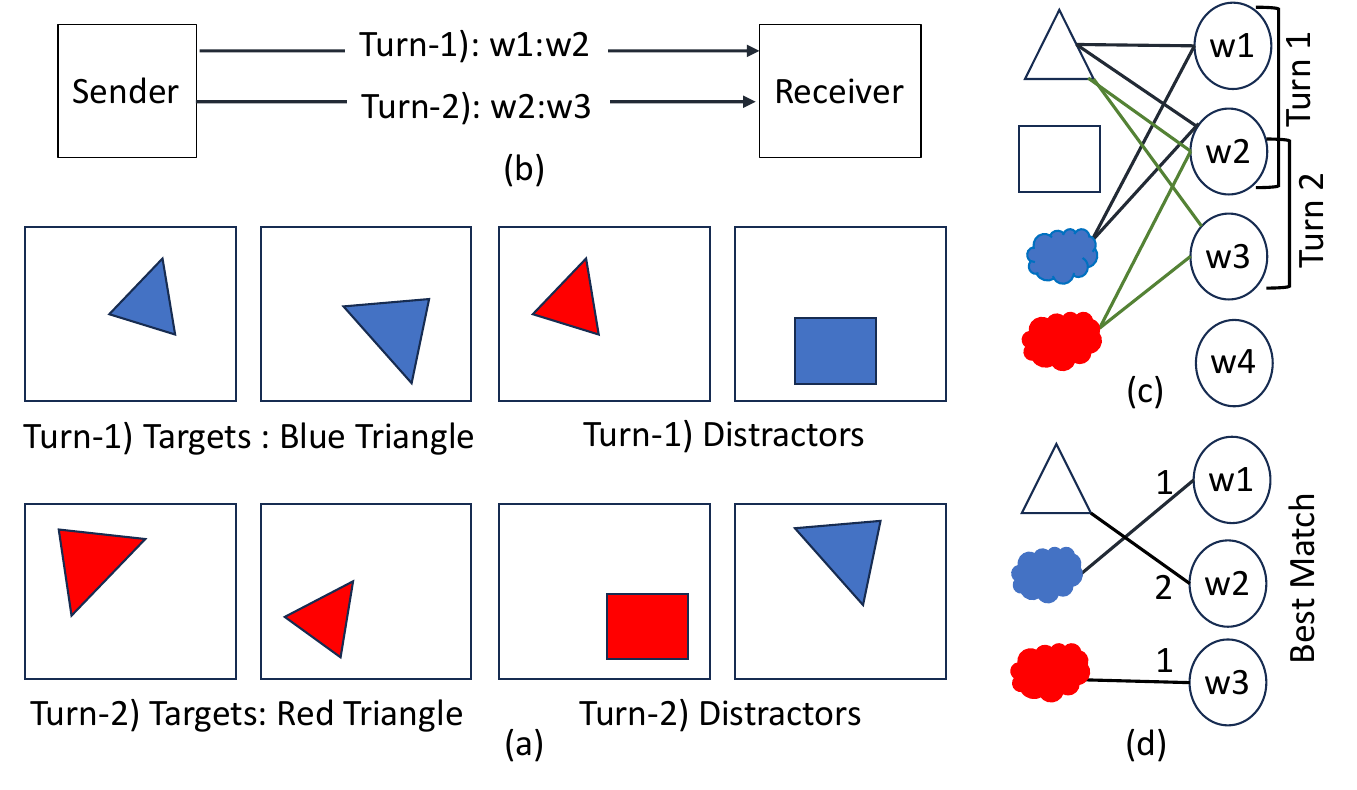}}
\caption{The multi-target shape game. 
(a) At each turn, the sender is given  a set of images, a subset of them marked as targets by an Oracle. (b) Sender generates messages $[w_1, w_2]$ for the \texttt{blue-triangle} (turn 1) and $[w_2, w_3]$ for the \texttt{red-triangle} (turn 2). (c) During evaluation, we construct a bipartite graph of words generated by the sender and concepts provided by the Oracle for each turn. (d) The best-match algorithm matches EC words to NL concepts and provides the CBM score. In this example, all EC words are matched with NL concepts, resulting in a CBM score of $1.0$.}
\label{fig:shape-game}
\end{figure}

Chief among these is compositionality, which enables the construction of complex meanings from the meaning of parts  \cite{li2019ease}. 
However, quantifying compositionality in EC is notoriously difficult \cite{lazaridou2018emergence}.
In fact, recent studies found that common compositionality measures in EC, e.g., topographic similarity \cite{brighton2006understanding,li2019ease}, do not correlate well with success in the task \citep{chaabouni2021emergent, yao2022linking}. 
Moreover, while seeking to assess an opaque protocol, common evaluation measures in EC are opaque themselves---they do not provide a human-interpretable characterization of the compositionality of the communication. 

This work develops a method that characterizes how compositional an EC is. Our method is founded on the key insight that, in an EC setting, a communication is compositional if agents communicate successfully via complex messages formed of simple atoms \emph{and} these atoms are mapped to NL concepts. 
%
Our evaluation is based on the classical best-match algorithm  \cite{hopcroft1973n}. Given a set of EC atoms and a set of NL concepts, we construct a bi-partite graph and seek the optimal one-to-one mapping between words and concepts. A perfectly compositional EC would yield a perfect match, where every EC atom is mapped to exactly one NL concept. 
See Figure \ref{fig:shape-game} for an illustration.

We experiment with the proposed procedure in two EC settings and compare it to two popular evaluation measures. 
We demonstrate that our approach provides a fine-grained characterization of emergent protocols, exposing their strengths and weaknesses, while other measures only provide coarse, opaque scores. According to our evaluation, state-of-the-art communication methods do not achieve satisfactory results.

\section{Background}
\subsection{Emergent Communication Setup} \label{sec:setup}
We follow a recent setup suggested by \citet{mu2021emergent}, where a sender needs to communicate with a receiver about a \emph{set of target objects} out of a larger set of candidate objects.
The sender sends a message to the receiver, which uses it to distinguish targets from distractor objects.
For example, to identify all red triangles out of objects with different shapes and colors. 
This setup is more conducive to emergence of compositional communication than the classical referential game \cite{lazaridou2016multi}.
Senders in this setup need to form a generalization rather than merely transmit the identity of a single target.


Formally, assume a world $\sX$, where each object is characterised by $n$ feature--value pairs (FVPs),  
$\langle f_1:v_1,\ldots,f_n:v_n \rangle $,  with feature $i$ having $k_i$  possible values, $v_i \in\{ f_i[1],\ldots,f_i[k_i]\}$. 
For instance, the Shape world \cite{kuhnle2017shapeworld} has objects like $\langle$shape:triangle,color:red$\rangle$. Labeling rules are boolean expressions over these FVPs.\footnote{In this work we only use conjunctive expressions.} 
Each rule $l : \sX \mapsto \{0,1\}$ labels each object as $0$ or $1$. For instance, 
the rule \texttt{Red Triangle} labels all red triangles as positive and all other objects as negative.
We identify the rule \texttt{Red Triangle} with the NL \emph{phrase} ``red triangle'', which comprises the \emph{concepts} ``red'' and ``triangle''.

At each turn of the game, we draw a set of candidate objects $\tilde{\sX} \subset \sX$ that is made of target objects, $\sT$, which obey the rule, and distractors, $\sD$, which do not: $l(x) = 1$ if  $ x \in \sT$ and $l(x) = 0 $ if $x \in \sD$. 

The sender encodes the set of target objects $\sT$ into a dense representation $\vu^s \in \sR^d$, and generates a message $m$, a sequence of \emph{words} from some vocabulary. 
The receiver encodes each candidate object $x \in \sX$ into a representation $\vu^r_x \in \sR^d$. It  decodes the message $m$ into a representation $\vz \in \sR^d$ and computes a matching score between each encoded candidate and the message representation, $g(\vz, \vu^r_x)$. 
A candidate is predicted as a target if its score is $>0.5$. 
The entire system---sender and receiver networks---is trained jointly with a binary cross entropy on correctly identifying each target. 

\subsection{Compositionality in EC}

Various measures have been proposed to evaluate if an EC protocol is compositional. We mention a few prominent ones below. 
However, while trying to capture the idea that a complex meaning uses the meaning of its parts, previous measures fail to provide a concrete \emph{mapping of parts}---EC words and NL concepts.
We say an emergent communication is compositional if the following conditions hold:
\begin{enumerate}[leftmargin=*,itemsep=1pt,parsep=1pt,topsep=1pt]
    \item EC words are mapped to NL concepts.
    \item A complex EC message is composed of simple EC words.
    \item Agents communicate successfully via complex messages.
    \item An EC message composed of words has the same meaning as an NL phrase composed of the respective concepts.
\end{enumerate}
Being able to compose EC words while preserving their NL meaning allows agents to generalize to new data while using interpretable communication.

\subsection{ Compositionality Evaluations in EC} \label{sec:top-ami}

We briefly describe two notable measures of compositionality in EC. 
 Appendix \ref{app:measures} gives more details. 

\vspace{5pt}
\noindent\textbf{Topographic Similarity (TopSim)} \cite{brighton2006understanding} measures how well messages align with object representations. 
Given a set of objects and their EC messages, 
calculate a matrix of distances between every two messages and a separate matrix of distances between every two object representations.  
TopSim is the Spearman 
correlation of the two distance matrices. 

TopSim is a global metric that does not require a reference language and can be applied to any EC setup. 
However, 
recent studies found that TopSim does not correlate well with success in the task being played \citep{chaabouni2021emergent, yao2022linking}.
More importantly, it does not provide a mapping between EC words and NL concepts, and thus cannot directly assess compositionality. 

\noindent\textbf{Adjusted Mutual Information (AMI)}
\cite{vinh2009information} 
measures the MIT 
between a set of EC messages and a corresponding set of NL phrases, adjusted for chance.
\citet{mu2021emergent} showed  AMI is a better compositionality measure 
than TopSim, as it directly assesses the MI between a message and its  NL 
phrases. 
Still, AMI operates at the level of messages and phrases, rather than atomic words and concepts, respectively.

\section{Concept Best Matching}

Our key insight is that compositionality requires a mapping between words (atomic parts of EC messages) and concepts (a set of FVPs) that compose NL phrases. 
Given an evaluation set of examples $D$, consider their corresponding EC messages $\sM$ and NL phrases $\sL$. 
Let $\sW$ denote the set of unique words in $\sM$, and $\sV$ the set of unique FVPs in $\sL$. Let $\{w\}_i$, $\{v\}_i$ be the sets of words and FVPs in sample $i$, and $m_i= |\{w\}_i|, l_i = |\{v\}_i|$ their sizes.
We construct a weighted bipartite graph $\gG = ((\sW, \sV), \mathbb{E})$ with words on one side and FVPs on the other side. 
Edges $\mathbb{E}$ are defined by the evaluation set. 
The weight $q_e$ of edge $e_{ij} $ is the number of times word $i$ appeared in a message that the sender transmitted for a labeling rule with FVP $j$.
This construction reflects the intuition that we do not know the correct mapping between words and FVPs (concepts). 
See Figure \ref{fig:shape-game} for an example. 

\vspace{-1pt}
\subsection{Best Match Algorithm}
\vspace{-1pt}

Given the graph $\gG$, we seek an optimal pairing between words and concepts (FVPs), such that no two edges share the same word nor the same concept.  
The score of a match $ \sB \subseteq \mathbb{E}$ is the sum of its edge weights, $\sum_{e \in \sB} q_e$. 
The best match, $\mathrm{BM} = \mathrm{BM}(\gG)$, maximizes this score:
\begin{equation}
\vspace{-2pt}
\label{eq:best_match}
   \text{BM} =  \argmax_{\sB} \sum_{e \in \sB}q_e  \hspace{.5cm} 
\end{equation}
Such an ideal mapping is fully interpretable. A high  score indicates that the agents learned to generate a unique EC word for each NL concept.

Normalizing the BM score by the number of symbols and words, $Q=\sum_{i \in D}\max(|m_i|, |l_i|)$, guarantees that $\mathrm{BM}\in [0, 1]$.\footnote{In practice, the lowest bound is  $q_{\tilde{e}} / Q $, where ${\tilde{e}} $ is the edge with the highest weight in $\gG $.}
The best match for a weighted bipartite graph can be found efficiently by the Hungarian algorithm \cite{kuhn1955hungarian,hopcroft1973n}.

\section{Experimental Setup}

\begin{table*}
\centering
\begin{adjustbox}{max width=\textwidth}
\begin{tabular}{l c c c c c c c c c c c c c r r r}
\toprule 
& & & & \multicolumn{2}{c}{NL} & \multicolumn{2}{c}{EC} & & \multirow{2}{*}[-8pt]{\shortstack{Top\\Sim}} & & \multicolumn{4}{c}{Best Match} \\ 
\cmidrule(lr){5-6} \cmidrule(lr){7-8} \cmidrule(lr){12-15} 
 &  Com  &  $l$  &  $d$   &  Cons  &  Phrs  &  $\#w$   &  $\#m$   &  Acc  &    &  AMI  &  CBM  &  Amb  &  Para  &  Unm\\
\midrule  
\multirow{4}{*}{\rotatebox[origin=c]{90}{Shape}}& GS & 1 & 17 & 17 & 17 & 11 & 11 & 0.84 & 0.44 & 0.76 & 0.62 & 0.37 & 0.01 & 0.25\\
& QT & 1 & 5 & 17 & 17 & 26 & 26 & 0.87 & 0.39 & 0.85 & 0.52 & 0.41 & 0.08 & 0\\
& GS & 4 & 17 & 17 & 34 & 16 & 226 & 0.78 & 0.34 & 0.38 & 0.52 & 0.48 & 0 & 0.01\\
& QT & 4 & 5 & 17 & 34 & 26 & 187 & 0.83 & 0.23 & 0.37 & 0.55 & 0.37 & 0.08 & 0\\
\midrule  
\multirow{4}{*}{\rotatebox[origin=c]{90}{Thing}} & GS & 1 & 50 & 50 & 50 & 18 & 18 & 0.86 & 0.45 & 0.72 & 0.38 & 0.62 & 0 & 0.55\\
& QT & 1 & 6 & 50 & 50 & 50 & 50 & 0.92 & 0.07 & 0.74 & 0.73 & 0.25 & 0.02 & 0\\
& GS & 4 & 50 & 50 & 901 & 36 & 198 & 0.63 & 0.33 & 0.03 & 0.24 & 0.76 & 0 & 0.28\\
& QT & 4 & 6 & 50 & 901 & 64 & 450 & 0.77 & 0.14 & 0.05 & 0.23 & 0.69 & 0.08 & 0\\
\bottomrule     
\end{tabular}
\end{adjustbox}
\caption{Results for \Shape{} and \Thing{} games with  Gumbel-Softmax (GS) or Quantized (QT) communication using rules of length $l$ and words of length $d$, showing number of unique NL concepts (Cons) and phrases (Phrs) and EC words ($\#w$) and messages ($\#m$).
TopSim and AMI give only coarse scores. Our method gives the best match score (CBM) and rates of Ambiguities (Amb), Paraphrases (Para), and Unmatched concepts (Unm).  
}
\label{tbl:games_main}
\vspace{-3pt}
\end{table*}


We experiments with agents that learn to play multi-target referential games, introduced by  \citet{mu2021emergent} and described in Section~\ref{sec:setup}.
All experiments reported in Table \ref{tbl:games_main} have the number of possible words $\geq$ the number of concepts, in each game, to allow perfect match of words to concepts.
In Appendix \ref{app:suboptimal_configs} we report more experiments with sub-optimal configurations.


\paragraph{Datasets.}
We experiment with two datasets, described briefly here; Appendix \ref{app:datasets} has more details. \\ 
\noindent  \textbf{(1)} \Shape{}: A visual reasoning dataset \cite{kuhnle2017shapeworld} of objects over a black background.
Our version of the dataset contains four attributes: 
shape, color, and horizontal and vertical positions. 
Overall, the \Shape{} dataset contains 17 concepts, and a maximal labeling rule of 4 FVPs. \\
\noindent \textbf{(2)} \Thing{}: A synthetic dataset of 100,000 objects. Each object has five attributes, each with 10 possible values.
Overall, the \Thing{} datasets contains 50 concepts and a maximal labeling rule of 5 FVPs.

\paragraph{Communication Channel.}
We train the system with two types of communication channels: the popular Gumbel-Softmax (GS) \cite{havrylov2017emergence,jang2016categorical} and quantized communication (QT) \cite{carmeli2022emergent}.  
In GS, each word is a $d$-dimensional one-hot vector. In QT, each word is a $d$-dimensional binary vector.
QT was shown to be superior and easier to optimzie compared to GS in EC games. 
In both cases we use a recurrent network to generate multi-word messages.
 Appendix \ref{app:impl_details} has implementation details.



\newcommand{\mycenterline}{%
\tikz{%
\draw (0.06,0.2) -- (0.34,0.2);
}%
}

\newcommand{\mymiddleline}{%
\tikz{%
\draw (0.2,0) -- (0.2,0.4);
\draw (0.06,0.2) -- (0.34,0.2);
}%
}

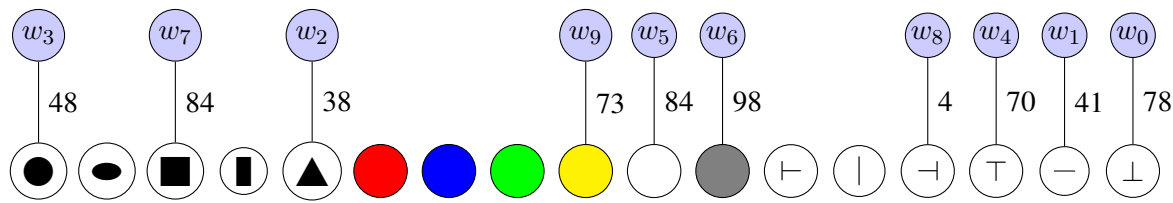
\begin{figure*}[t]
\centering
\begin{tikzpicture}[scale=.9]
\node[draw, circle, inner sep=2pt, fill=blue!20] (W1) at (0,0) {$w_3$};
  \node[draw, circle, inner sep=2pt, fill=blue!20] (W3) at (2,0) {$w_7$};
 \node[draw, circle, inner sep=2pt, fill=blue!20] (W5) at (4,0) {$w_2$};
  \node[draw, circle, inner sep=2pt, fill=blue!20] (W9) at (8,0) {$w_{9}$};
  \node[draw, circle, inner sep=1pt, fill=blue!20] (W10) at (9,0) {$w_{5}$};
  \node[draw, circle, inner sep=1pt, fill=blue!20] (W11) at (10,0) {$w_{6}$};
  \node[draw, circle, inner sep=1pt, fill=blue!20] (W14) at (13,0) {$w_{8}$};
  \node[draw, circle, inner sep=1pt, fill=blue!20] (W15) at (14,0) {$w_{4}$};
  \node[draw, circle, inner sep=1pt, fill=blue!20] (W16) at (15,0) {$w_{1}$};
  \node[draw, circle, inner sep=1pt, fill=blue!20] (W17) at (16,0) {$w_{0}$};

  \node[draw, circle, inner sep=2pt] (circle) at (0,-2) {\CircleSolid};
  \node[draw, circle, inner sep=2pt] (ellipse) at (1,-2) {\EllipseSolid};
  \node[draw, circle, inner sep=2pt] (square) at (2,-2) {\SquareSolid};
  \node[draw, circle, inner sep=2pt] (rectangle) at (3,-2) {\RectangleBold};
  \node[draw, circle, inner sep=2pt] (triangle) at (4,-2) {\TriangleUp};
  \node[draw, circle, inner sep=2pt, minimum size=20pt, fill=red] (red) at (5,-2) {};
  \node[draw, circle, inner sep=2pt, minimum size=20pt, fill=blue] (blue) at (6,-2) {};
  \node[draw, circle, inner sep=2pt,  minimum size=20pt, fill=green] (green) at (7,-2) {};
  \node[draw, circle, inner sep=2pt,  minimum size=20pt, fill=yellow] (yellow) at (8,-2) {};
  \node[draw, circle, inner sep=2pt,  minimum size=20pt, fill=white] (white) at (9,-2) {};
  \node[draw, circle, inner sep=2pt,  minimum size=20pt, fill=gray] (gray) at (10,-2) {};
  \node[draw, circle, inner sep=3pt] (left) at (11,-2) {\rotatebox[origin=c]{90}{$\top$}};
  \node[draw, circle, inner sep=3pt] (middle) at (12,-2) {$\mid$};
  \node[draw, circle, inner sep=3pt] (right) at (13,-2) {\rotatebox[origin=c]{90}{$\bot$}};
  \node[draw, circle, inner sep=3pt] (top) at (14,-2) {$\top$};
  \node[draw, circle, inner sep=4pt] (center) at (15,-2) {\mycenterline{}};
  \node[draw, circle, inner sep=3pt] (bottom) at (16,-2) {$\bot$};

\draw[color=black] (W1) -- node[right] {\color{black}48} (circle);
\draw[color=black] (W3) -- node[right] {\color{black}84} (square);
\draw[color=black] (W5) -- node[right] {\color{black}38} (triangle);
\draw[color=black] (W9) -- node[right] {\color{black}73} (yellow);
\draw[color=black] (W10) -- node[right] {\color{black}84} (white);
\draw[color=black] (W11) -- node[right] {\color{black}98} (gray);
 \draw[color=black] (W14) -- node[right] {\color{black}4} (right);
 \draw[color=black] (W15) -- node[right] {\color{black}70} (top);
\draw[color=black] (W16) -- node[right] {\color{black}41} (center);
\draw[color=black] (W17) -- node[right] {\color{black}78} (bottom);

\end{tikzpicture}
\caption{The $ \text{word} \leftrightarrow \text{FVP} $ best-match graph for the \Shape{} game (GS communication, $l=1$). 
The algorithm matched just $10$ concepts to EC words out of $17$ possible concepts. 
}
\label{fig:sbm-graph}
\vspace{-3pt}
\end{figure*}

\section{Results} \label{sec:results}

Table~\ref{tbl:games_main} shows  results on several configurations of the two games.
As expected, accuracy degrades as the task becomes more difficult, with longer labeling rules. QT communication yields better task accuracy than GS, consistent with \citet{carmeli2022emergent}. 
TopSim is not well correlated with accuracy, as found in prior work (Section \ref{sec:top-ami}). 
AMI and CBM are more correlated with accuracy. 
With long messages and labeling rules ($l=4$), AMI is less correlated with accuracy. This may be explained by AMI operating at the level of whole messages and labeling rules---AMI struggles with large numbers of unique messages and NL phrases. 
In contrast, CBM assesses word-to-concept matching, so it is less affected by long messages and rules.
However, AMI and CBM are well correlated (Pearson $0.7$ and $0.79$ on the \Shape{} and \Thing{} games).

The table also shows the rate of ambiguities and paraphrases, exposed by our method.  
\textbf{Ambiguities} happen when the same EC word is mapped to different concepts.
 Table \ref{tbl:games_main} shows higher rates of ambiguities in GS compared to QT communication, and in longer rules. 
\textbf{Paraphrases} are EC words that do not have a best-matched concept node. 
Paraphrases occur more with QT, which enables 32 and 64 unique EC words for the \Shape{} and \Thing{} games, while these games have only 17 and 50 NL concepts, respectively.
GS allows setting the number of unique words equal to number of concepts, thus is less exposed to this sub-optimal phenomenon.

Finally, 
\textbf{Unmatched concepts} have no matching EC word. They result from a too narrow or underutilized channel. 
As  Table \ref{tbl:games_main} shows, GS communication suffers from a high rate of unmatched concepts (up to  $0.5$). This phenomenon can be observed by the low \# of unique words (18) compared to \# of unique concepts (50) for the \Thing{} game with GS and $l=1$. QT has a sufficient number of words and thus no unmatched concepts.

\paragraph{Example Match.}
Beyond global scores, the CBM provides an interpretable \emph{translation graph}, which maps EC words to NL concepts, facilitating insights on reasons for sub-optimal communication. 
%
Figure \ref{fig:sbm-graph} shows an example graph for the \Shape{} game with GS communication.\footnote{The exact configuration is given in Table \ref{tbl:games_main}, first row.} As seen, the algorithm successfully matched $10$ EC words to $10$ concepts. The sender generated 11 unique words in this experiment,  
so 
one word do not matched to any concept even in this narrow-channel setup. 
\section{Conclusion}

We proposed a new procedure for assessing compositionality of emergent communication.
In contrast to other evaluations, our procedure provides a human-interpretable translation table of emerged words to natural concepts. Moreover, our approach provides detailed insights into the reasons for sub-optimal translation. We demonstrated it on two EC games with two different communication types. Our evaluation revels even that quantized communication performs better than Gumbel-softmax, yet none exhibits compositionality in a level similar to natural language.

\clearpage 

\pagebreak
\section{Limitations}
Our analysis is limited to datasets where gold label phrases exist.
The evaluated dataset should be composable and the gold language should use finite set of concepts.
Preferably, these concepts can be classified into several categories.
We further assume an Orcal function that can divide the objects to targets an distractors during training.
Importantly, our approach does not require labels during training.
\bibliography{anthology,references}
\bibliographystyle{acl_natbib}
\clearpage 

\pagebreak
\appendix
\vspace{-24pt}
\textbf{{\Large Appendices}}
\vspace{24pt}
\section{Details on Evaluation Measures} \label{app:measures}

We provide here information on how to calculate the topographic similarity (TopSim) \cite{brighton2006understanding, lazaridou2018emergence, yao2022linking} and adjusted mutual information (AMI) measures \cite{vinh2009information, mu2021emergent}, in the context of EC. Refer to the original papers for full details. 

\paragraph{Topographic Similarity.} The topographic similarity measures how messages align with the object representations. 
Concretely, let $ \text{cos}_{ij} = \text{cos}(\vx_i,\vx_j) = \vx_i \cdot \vx_j / (||\vx_i||_2 ||\vx_j||_2) $ 
be the cosine similarity of object representations $\vx_i$ and $\vx_j$, and $\text{edit}_{ij} = \text{edit}(m_i, m_j)$ be
the Levenshtein distance \cite{levenshtein1966binary} between messages $m_i$ and $m_j$. 
Let $\text{ncos} = -\{\cos_{ij}\}_{ij}$ be the list of negative cosine similarities,  
$\text{edit} = \{\text{edit}_{ij}\}_{ij}$ the list of Levenshtein distances, 
and $R(\cdot)$ the ranking function. 
Then the topographic similarity is the Spearman rank correlation of the two matrices:
\begin{equation} \label{eq:top_sim}
    \mathrm{Topsim} = \rho(R(\text{edit}),R(\text{ncos}))
\end{equation}
where $\rho$ is the standard Pearson correlation. 

Topographic similarity is a global metric that does not require a reference language and can be easily applied to every EC setup. 
However, despite its popularity, recent studies found that topographic similarity does not correlate well with success in the task being played  \citep{chaabouni2021emergent, yao2022linking}.
More important, it does not provide a mapping between EC atoms and NL concepts, and thus cannot directly assess compositionality. 

\paragraph{Adjusted Mutual Information.}
Given a test set $D$, let $\sM$ denote the set of messages generated by the sender and $\sL$ denote the set of NL phrases (e.g., \texttt{Red Triangle}, \texttt{Blue Square}) exist as labels in the data. 
The adjusted mutual information (AMI) \cite{vinh2009information} measures the mutual information between messages and labels, adjusted for chance:
\begin{equation} \label{eq:ami}
\small
    \text{AMI}(M, L) = \frac{I(M,L) - \mathbb{E}(I(M,L)}{\max(H(M), H(L)) - \mathbb{E}(I(M, L))}
\end{equation}
where $I(M, L)$ is the mutual information between $M$ and $L$, $H(\cdot)$ is the entropy, and $\mathbb{E}(I(M, L))$ is computed with respect to a hypergeometric model of randomness.

\begin{table*}
\centering
\begin{tabular}{l c c c c c c c c }
\toprule 
Batch &     &   Num      & Num  & Cell & Sender  & Sender  & Receiver & Receiver \\
Size & lr   &   Targets  & Distr & Type & Hidden  & Embed  & Hidden & Embed \\
\midrule 
10 & 0.0005 & 8/20 & 8/20 & LSTM & 60 & 60 & 40 & 40\\
\bottomrule     
\end{tabular}
\caption{Hyper-parameters for the agents in the \Shape{} game.}
\label{tbl:impl_params}
\end{table*}

\begin{table*}
\small
\centering
\begin{tabular}{l c c c c c c c c c}
\toprule 
Exp &   & Comm & Label & word & Msg &  Best  &      &   & Best\\
Num & Game& Type & Len   & Len  & Len & Epoch & ACC  & AMI& Match\\
\midrule 
1 & \Shape{} & gs & 1 & 8 & 4 & 97 & 0.829 & 0.404 & 0.405\\
2 & \Shape{} & gs & 1 & 64 & 4 & 96 & 0.843 & 0.390 & 0.339\\
\midrule 
3 & \Shape{}& gs & 4 & 8 & 1 & 8 & 0.742 & 0.490 & 0.152\\
4 & \Shape{}& gs & 4 & 64 & 1 & 10 & 0.748 & 0.413 & 0.130\\
\midrule 
5 & \Shape{}& qt & 1 & 3 & 4 & 86 & 0.878 & 0.540 & 1.401\\
6 & \Shape{}& qt & 1 & 64 & 4 & 85 & 0.991 & 0.000 & 0.001\\
\midrule 
7 & \Shape{}& qt & 4 & 3 & 1 & 9 & 0.768 & 0.474 & 0.152\\
8 & \Shape{}& qt & 4 & 64 & 1 & 12 & 0.897 & 0.021 & 0.009\\
\midrule 
9 & \Thing{}& gs & 1 & 8 & 4 & 83 & 0.793 & 0.102 & 0.091\\
10 & \Thing{}& gs & 1 & 64 & 4 & 94 & 0.798 & 0.074 & 0.151\\
\midrule 
11 & \Thing{}& gs & 4 & 8 & 1 & 97 & 0.655 & 0.016 & 0.052\\
12 & \Thing{}& gs & 4 & 64 & 1 & 99 & 0.626 & 0.014 & 0.058\\
\midrule 
13 & \Thing{}& qt & 1 & 3 & 4 & 96 & 0.838 & 0.267 & 0.119\\
14 & \Thing{}& qt & 1 & 64 & 4 & 100 & 0.973 & 0.000 & 0.014\\
\midrule 
15 & \Thing{}& qt & 4 & 3 & 1 & 26 & 0.622 & 0.020 & 0.057\\
16 & \Thing{}& qt & 4 & 64 & 1 & 96 & 0.844 & 0.000 & 0.013\\
\bottomrule     
\end{tabular}
\caption{Configurations for more \Shape{} and \Thing{} game experiments. In these configuration we intentionally define sub-optimal communication parameters in order to demonstrate the usefulness of the evaluation metrics. Experiment results are provided in Table \ref{tbl:suboptimal_results}}
\label{tbl:suboptimal_config}
\end{table*}

\begin{table*}
\small
\centering
\begin{tabular}{l r r r c c r r r r r r r r r r  }
\toprule 
Exp & Unq & Unq & Unq  & Ung & Totl & Good  & Amb  & Phrs & Totl & Unm & Par &    & \\
Num & Msgs & Wrds & Prs  & Cons & Edgs & Edgs  & Edgs  & Edgs  & Cons  & Cons & Scor & Prc & Rcl\\
\midrule 
1& 258 & 7 & 17 & 17 & 4k & 1621 & 2379 & 0 & 1k & 444 &  0 & 0.137 & 0.546\\
2 & 442 & 55 & 17 & 17 & 4k & 1356 & 1724 & 920 & 1k & 0 & 0.23 & 0.144 & 0.577\\
\midrule
3 & 8 & 8 & 34 & 17 & 4k & 610 & 390 & 0 & 4k & 1457 & 0 & 0.61 & 0.152\\
4& 44 & 44 & 34 & 17 & 4k & 520 & 326 & 154 & 4k & 0 &0.04 & 0.52 & 0.13\\
\midrule
5  & 238 & 8 & 17 & 17 & 4k & 1605 & 2395 & 0 & 1k & 385 &0 & 0.144 & 0.577\\
6 & 1000 & 3899 & 17 & 17 & 4k & 46 & 0 & 3954 & 1k & 0 &  0.99 & 0.011 & 0.044\\
\midrule
7 & 7 & 7 & 34 & 17 & 4k & 607 & 393 & 0 & 4k & 1727 &  0 & 0.607 & 0.152\\
8 & 974 & 974 & 34 & 17 & 4k & 38 & 0 & 962 & 4k & 0 &  0.24 & 0.038 & 0.009\\
 \midrule
9 & 543 & 7 & 50 & 50 & 4k & 366 & 3634 & 0 & 1k & 807 &  0 & 0.048 & 0.19\\
10 & 806 & 60 & 50 & 50 & 4k & 601 & 3369 & 30 & 1k & 0 &  0.01 & 0.1 & 0.399\\
\midrule
11 & 8 & 8 & 901 & 50 & 4k & 210 & 790 & 0 & 4k & 3332 &  0 & 0.21 & 0.052\\
12 & 21 & 21 & 901 & 50 & 4k & 233 & 767 & 0 & 4k & 2267 & 0 & 0.233 & 0.058\\
\midrule
13 & 412 & 8 & 50 & 50 & 4k & 478 & 3522 & 0 & 1k & 795 &  0 & 0.046 & 0.184\\
14 & 1000 & 3991 & 50 & 50 & 4k & 57 & 0 & 3943 & 1k & 0 & 0.99 & 0.013 & 0.052\\
\midrule
15 & 11 & 6 & 901 & 50 & 4k & 229 & 771 & 0 & 4k & 3483 &  0 & 0.229 & 0.057\\
16 & 1000 & 1000 & 901 & 50 & 4k & 50 & 0 & 950 & 4k & 0 & 0.24 & 0.05 & 0.013\\
\bottomrule     
\end{tabular}
\caption{Results for more \Shape{} and \Thing{} game configurations described in Table \ref{tbl:suboptimal_config}. In these experiments we intentionally define sub-optimal communication parameters in order to demonstrate the usefulness of the evaluation metrics.}
\label{tbl:suboptimal_results}
\end{table*}

\section{Datasets}
\label{app:datasets}
\begin{figure}
\centering
\resizebox{6cm}{!} {\includegraphics{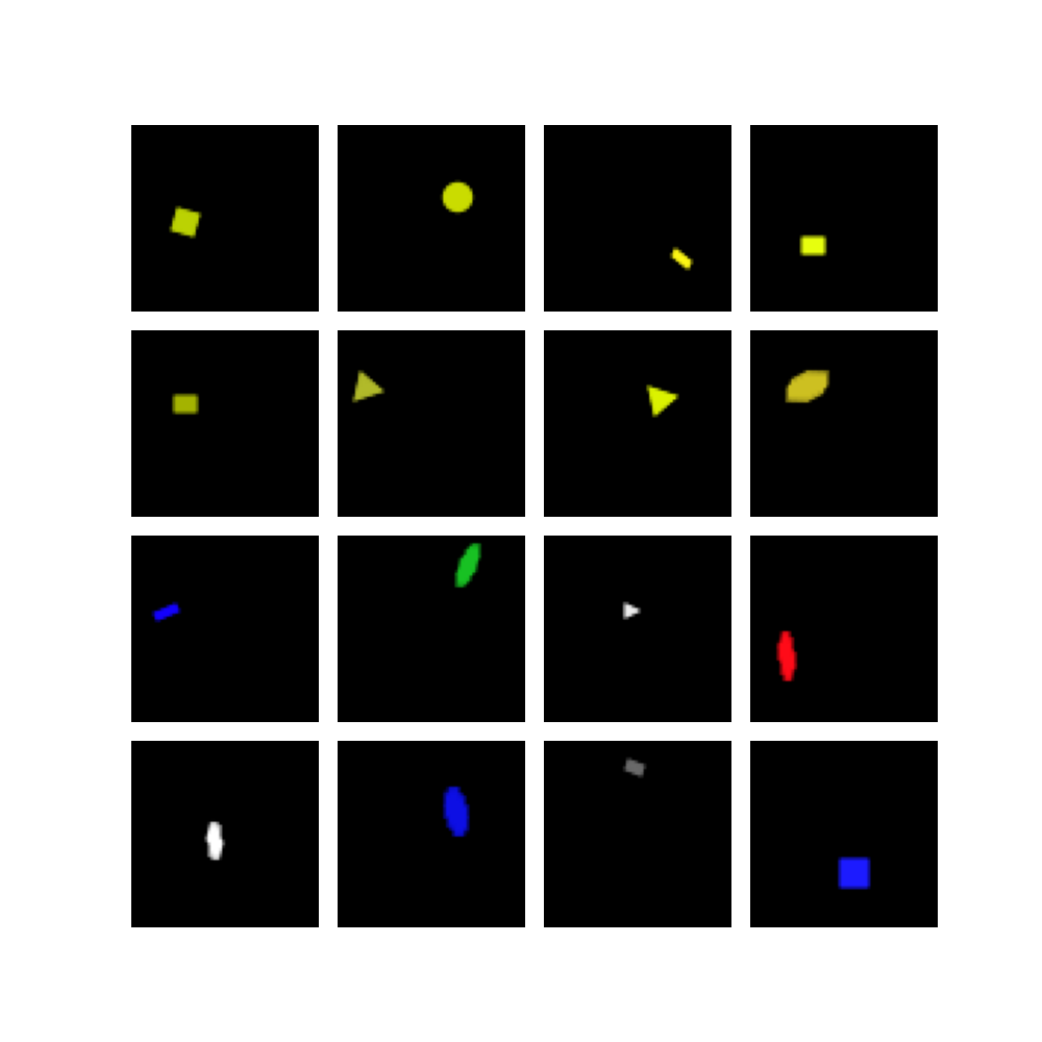}}
\caption{The \Shape{} dataset, presenting one turn. Top $8$ images are \texttt{Yellow} targets. Bottom $8$ images are distractors. 
}
\vspace{-10pt}
\label{fig:shape_dataset}
\end{figure}

We use two datasets in our experiments. Both will be made publicly available. 
\paragraph{Shape.} A visual reasoning dataset \cite{kuhnle2017shapeworld} of objects over a black background.
Our version of the dataset contains four attributes: the shape attribute has five shapes, the color attribute has six colors. To these we added horizontal and vertical position attributes each with three values.
Overall, the \Shape{} dataset contains 17 concepts, and a maximal labeling rule of 4 FVPs which sum to 270 unique 4-concept labels. The dataset contains many images that obey the same label. See Figure \ref{fig:shape_dataset} for an example. 

\paragraph{Thing.} A synthetic dataset of 100,000 objects which we design for controlled experiments. Each object in the dataset has five attributes, each with 10 possible values.
Overall, the \Thing{} datasets contains 50 concepts and a maximal labeling rule of 5 FVPs.
We experiments with label lengths varied from $1$, describing a single concept, to $5$, which completely describes an object in the dataset.

\section{Implementation Details}
\label{app:impl_details}
We run all experiments over a modified version of the Egg framework \cite{kharitonov2019egg_a20}.\footnote{Our modification will be released as open-source upon de-anonymization.}
In our version the communication modules used by the sender and the receiver are totally separated from their perceptual modules, which we term `agents'.
The Quantized (QT) and Gumbel-Softmax (GS) protocols are implemented at the communication layer.
For GS we use an implementation provided by the Egg framework, where we do not allow temperature to be learned, and set the straight-through estimator to False.
For QT we followed parameter recommendations from \citet{carmeli2022emergent} and use a binary quantization in all experiments.
Both GS and QT uses a recurrent neural network (RNN) for generating multiple words in a message. See Table \ref{tbl:impl_params} for RNN details.

The agents for the \Thing{} game use fully connected feed-forward networks with input dimension of $d=270$.
Objects are encoded by concatenating five one-hot vectors, one per attribute. There are 50 unique values and 4 communication words (SOS, EOS, PAD, UNK), thus the total object representation length is $d=270$ 
The agents of the \Shape{} game are implemented with a convolutional neural network similar to \citet{mu2021emergent}.
Network hyper-parameters are provided in Table \ref{tbl:impl_params}. 
In the experiments, we report results when varying the communication elements (word length and message length).
Experiments in  Table \ref{tbl:games_main} are done with 20 targets and 20 distractors, while experiments reported in Tables \ref{tbl:suboptimal_config} and \ref{tbl:suboptimal_results} are done with 8 targets and 8 distractors.
Our main interest in this work is to evaluate the emerged communication and not to achieve the best possible performance, thus we did not conduct a thorough hyper-parameter search for the agents themselves.

We run all experiments on a single A100 GPU with 40 GB of RAM. Model size is less than 10M.
We usually trained the model for 200 epochs which took about 10 hours to complete.
Our code will be made publicly available upon de-anonymization. 

\section{Evaluating Sub-optimal Configurations}
\label{app:suboptimal_configs}
Here we provide results from experiments with several sub-optimal configurations and demonstrate how CBM identifies these deficiencies. 
We report configuration parameters for each experiment together with accuracy, AMI, and CBM Score in Table \ref{tbl:suboptimal_config}.
We report all CBM metrics for these experiments in Table \ref{tbl:suboptimal_results}.
Information from the two tables is aligned by the experiment number (first column).

Sub-optimal communication may stem from a channel that is either too wide or too narrow.
Narrow channels are restricted by a small number of words and short messages.
Wide channels allow long words and/or long messages.
Past studies \cite{slonim2002information, tucker2022generalization} suggest that creating an information bottleneck by narrowing the channel encourages generalization.
As seen, CBM scores for most reported sub-optimal configurations are low.
Results from experiments 1, 2, and 5 are the only ones for which the CBM score is higher than $0.2$. 
Experiment 6 (\Shape{}) achieves almost perfect accuracy (0.991) due to its wide channel. However, the paraphrase score for this experiment is $0.989$ showing that the sender uses many different words to refer to the same concept. This phenomenon can also be seen by the large number of unique words ($3899$) generated for just $17$ unique concepts.
A similar phenomenon is observed for the \Thing{} game (experiment 14).

Experiment 11 (\Thing{}) uses a narrow channel. The vocabulary contains only 8 words and the channel allows one-word-long messages.
This configuration resulted in $3332$ concepts that do not match any word.
Interestingly, the number of unmatched concepts remains high even when extending the word length to $64$, in experiment 12. We attribute this sub-optimal phenomenon to GS communication deficiencies. The number of unique words generated by the sender in this experiment is only $21$ out of the possible 64 words allowed by the configuration.
For the \Shape{} game, on the other hand, the same wide-channel configuration resulted in $44$ unique words generated by the sender for the 17 concepts available in this game and all concepts have matched words.
\begin{figure}[t]
\centering
\resizebox{7cm}{!} {\includegraphics{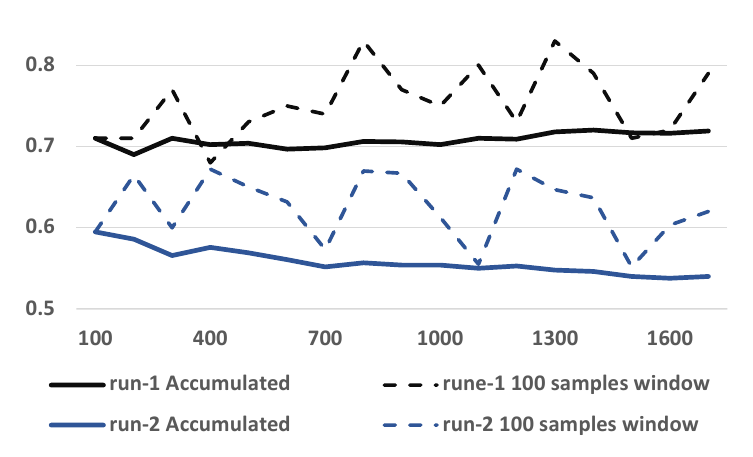}}
\caption{Assessing CBM sensitivity to the size of the evaluation set.
We show results for two randomly selected experiments.
As seen, the CBM score stabilizes when assessing datasets comprising more than $500$ samples.
}
\label{fig:data-sensitivity}
\end{figure}

\paragraph{Precision and recall.} After finding the best match, precision and recall between words and concepts can be computed.
We define message precision (Prc) and recall (Rcl) to be the number of matched edges in a message, divided by the number of words or concepts, respectively.
Precision and recall add insights in situation where the number of words in messages differs significantly from the number of concepts used for labeling the data.
Experiment 5 (\Shape{}) shows the highest recall, gained partially due to the long message length (four words) allowed by the channel, compared to the label length (one concept) dictated by the data.
In contrast, experiment 3 (\Shape{}) shows the highest precision. Indeed, this configuration restricts message the length to one word while the data dictates four-concept long phrases.

\section{CBM Sensitivity to  Dataset Size}
We evaluated CBM scores on increased data sizes.
Figure \ref{fig:data-sensitivity} shows results for two randomly selected experiments.
The solid lines indicate the CBM score, computed when accumulating data samples in 100 quintets.
The dashed lines indicate CBM score calculated independently for 18 successive data segments of $100$ samples each.
As seen, scores stabilize for evaluation sets of size larger than $500$ samples.
Interestingly, the $100$-sample scores (dashed) lines are higher than the accumulated scores (solid lines).
This is in line with our insight that low evaluation sizes result in higher match scores.
Standard deviation for the $18$ accumulated measures is $0.009$ and $0.016$, for the two experiments, respectively.
Standard deviation for the $18$ segmented measures is $0.044$ and $0.040$, for the two experiments, respectively.

\end{document}